\documentclass{article} 

\usepackage[dvipsnames]{xcolor}         
\definecolor{linkColor}{rgb}{0.18,0.39,0.62}
\definecolor{gray}{rgb}{0.5,0.5,0.5}
\definecolor{gray94}{gray}{.92}
\definecolor{gray90}{gray}{.90}
\definecolor{gray85}{gray}{.85}

\usepackage{iclr2025_conference,times}

\usepackage{amsmath,amsfonts,bm}

\def\eqref#1{equation~\ref{#1}}

\def\1{\bm{1}}

\DeclareMathAlphabet{\mathsfit}{\encodingdefault}{\sfdefault}{m}{sl}
\SetMathAlphabet{\mathsfit}{bold}{\encodingdefault}{\sfdefault}{bx}{n}

\usepackage[utf8]{inputenc} 
\usepackage[T1]{fontenc}    
\usepackage[colorlinks=true,linkcolor=linkColor,citecolor=linkColor,filecolor=linkColor,urlcolor=linkColor]{hyperref}       
\usepackage{url}            
\usepackage{booktabs}       
\usepackage{amsfonts}       
\usepackage{nicefrac}       
\usepackage{microtype}      
\usepackage{tcolorbox} 
\usepackage{graphicx}
\usepackage{arydshln}

\tcbuselibrary{breakable} 
\usepackage{pifont}
\usepackage{marvosym}
\usepackage{caption}
\usepackage{subcaption}
\usepackage{makecell}
\usepackage{csquotes}
\usepackage{epigraph}
\usepackage{amssymb}
\usepackage{mathtools}
\usepackage{amsthm}
\usepackage{algorithm}
\usepackage{algpseudocode}
\usepackage{algorithmicx}
\usepackage{amsmath}
\usepackage{multirow}
\usepackage{pifont}
\usepackage{mathtools}
\usepackage{enumitem}
\usepackage{setspace}
\usepackage{soul}
\usepackage{graphicx}
\usepackage{wrapfig}
\usepackage{makecell}

\usepackage[customcolors,norndcorners]{hf-tikz}

\newcommand\ourmethod{\textsc{TAPO}}
\newcommand\ourmethodfullname{\textsc{Textual Aesthetics Preference Optimization}}
\newcommand\ourdata{\textsc{TexAes}}
\newcommand\ourmodel{LLaMA-3.1-8B-TAPO}
\newcommand\ourlrgmodel{LLaMA-3.1-70B-TAPO}
\newcommand\ourbasemodel{LLaMA-3.1-8B-Instruct}
\newcommand\ourlrbasemodel{LLaMA-3.1-70B-Instruct}

\theoremstyle{plain}

\theoremstyle{definition}

\theoremstyle{remark}

\usepackage{wrapfig}
\usepackage{colortbl}
\usepackage{pifont}
\usepackage{wrapfig}
\usepackage{epstopdf}     
\usepackage{wrapfig}
\usepackage{cuted}
\usepackage{capt-of}

\title{Textual Aesthetics in Large Language Models}

\author{
Lingjie Jiang$^{1,2}$\thanks{Contribution during internship at Microsoft. \textsuperscript{\Letter} Corresponding Author.},\quad Shaohan Huang$^{1,}$\textsuperscript{\Letter},\quad Xun Wu$^1$,\quad Furu Wei$^1$ \\
$^1$Microsoft Research Asia \quad $^2$Peking University\\
\small{\texttt{lingjiejiang@stu.pku.edu.cn;} ~\texttt{\{shaohanh,~xunwu,~fuwei\}@microsoft.com}}\\
}

\iclrfinalcopy 
\begin{document}

\maketitle

\begin{abstract}
Image aesthetics is a crucial metric in the field of image generation. However, textual aesthetics has not been sufficiently explored. With the widespread application of large language models (LLMs), previous work has primarily focused on the correctness of content and the helpfulness of responses. Nonetheless, providing responses with textual aesthetics is also an important factor for LLMs, which can offer a cleaner layout and ensure greater consistency and coherence in content.
In this work, we introduce a pipeline for aesthetics polishing and help construct a textual aesthetics dataset named \ourdata{}. We propose a textual aesthetics-powered fine-tuning method based on direct preference optimization, termed \ourmethod{}, which leverages textual aesthetics without compromising content correctness. Additionally, we develop two evaluation methods for textual aesthetics based on text and image analysis, respectively.
Our experiments demonstrate that using textual aesthetics data and employing the \ourmethod{} fine-tuning method not only improves aesthetic scores but also enhances performance on general evaluation datasets such as AlpacalEval and Anera-hard. Our code and data are available at \url{https://github.com/JackLingjie/Textual-Aesthetics}.
\end{abstract}

\section{Introduction}
\label{sec:intro}


Image aesthetics~\citep{huang2024aesexpert,murray2012ava,kong2016photo,t1,c1} has emerged as a prominent research area within computer vision, focusing on assessing and improving the visual appeal of images. Aesthetics has recently been integrated into state-of-the-art image generation models, such as diffusion models~\citep{rombach2022high}, significantly enhancing the visual quality of generated images~\citep{wu2024q,wu2023q} and aligning them more closely with human preferences~\citep{huang2024aesexpert,wu2024multimodal,wu2023q}.

Meanwhile, advancements in large language models (LLMs) like ChatGPT~\citep{ChatGPT} and LLaMA~\citep{touvron2023llama2openfoundation,dubey2024llama3herdmodels} have demonstrated impressive generative capabilities across various domains, including code, articles, and web content. Although LLMs have made significant progress in generating textual content, enhancing the aesthetic quality of their output remains a critical challenge. A more aesthetically appealing and organized output not only improves user engagement by making the content more intuitive and comfortable to read but also enhances consistency and coherence. Consequently, exploring the textual aesthetics of LLMs is a highly desirable area of research.
%

In this work, we present the first investigation into improving the aesthetic quality of text generated by LLMs. Unlike image aesthetics benefiting from numerous large-scale aesthetic datasets (e.g., AVA~\citep{murray2012ava} and AesBench~\citep{huang2024aesbench}), advanced aesthetic learning technology~\citep{huang2024aesexpert, t2, t3, c3} and reliable aesthetic evaluation methods~\citep{deng2017image,su2011scenic}, textual aesthetics in LLMs lacks similar resources and established models.

To address this challenge, we first designed an aesthetic data generation pipeline leveraging GPT-4o for aesthetic polishing. This scalable pipeline can generate large volumes of high-quality aesthetic preference data. Based on this framework, we constructed the first aesthetic dataset in the LLM domain, \ourdata{}, which contains a total of 50,390 prompts data.

Based on \ourdata{}, existing post-training techniques such as DPO~\citep{dpo} can be used to fine-tune current LLMs at the aesthetic level. However, we found that directly applying these techniques not only failed to align effectively with the characteristics of our \ourdata{}, limiting its impact on aesthetic fine-tuning, but also negatively impacted the overall performance of these LLMs. To address this issue, we propose \textbf{T}extual \textbf{A}esthetics \textbf{P}reference \textbf{O}ptimization~(TAPO) which employs the Plackett-Luce~\citep{luce1959individual,plackett1975analysis} model with adjustable optimization weights to better leverage our dataset and enhance aesthetic fine-tuning performance. Furthermore, to better assess the aesthetic quality of LLM outputs, we have developed two evaluation pipelines: one based on text and the other based on images, respectively.

To validate the effectiveness of our \ourdata{} and \ourmethod{}, we performed aesthetic fine-tuning on the open-source LLaMA series models~\citep{dubey2024llama3herdmodels} and compared the aesthetic scores of the fine-tuned LLMs with state-of-the-art LLMs at different scales (from 8B to 70B). Additionally, to ensure objective and reliable results, we employed human experts for professional evaluation. Extensive experimental results ultimately demonstrated the effectiveness of our \ourdata{} and \ourmethod{}.

Our main contributions are listed as follow:
\vspace{-1mm}
\begin{itemize}[leftmargin=1.5em,itemsep=0pt,parsep=0.2em,topsep=0.0em,partopsep=0.0em]
    \item To the best of our knowledge, we for the first time indciate the crucial issue of exploring and improving the textual aesthetics in LLMs.
    \item We systematically identify the lack of related textual aesthetics datasets, and introduce a novel pipeline for aesthetic text polishing and contribute to the construction of a textual aesthetics dataset, named \ourdata{}.
    \item Based on \ourdata{}, we propose a DPO-based aesthetic fine-tuning algorithm, named \ourmethod{}, to effectively enhances the LLMs' aesthetic quality while preserving its general performance.
    \item Both qualitative and quantitative extensive experiments demonstrate that utilizing \ourdata{} and \ourmethod{} not only improves aesthetic scores but also enhances the general capabilities of LLMs. 
\end{itemize}

\section{Related Works}


\subsection{Image Aesthetics}
Image aesthetics~\citep{huang2024aesexpert,murray2012ava,kong2016photo} is a subfield of computer vision that focuses on assessing~\citep{deng2017image,su2011scenic} and improving the aesthetic quality of images~\citep{bhattacharya2010framework,deng2018aesthetic}. Early work in the field of image aesthetics focused on using handcrafted metrics to assess aesthetic scores~\citep{nack2001computational,neumann2005defining}. However, with the development of deep learning, there has been significant interest in applying CNN~\citep{c1,c2,c3} or Transformer~\citep{t1,t2,t3,t4} based methods to solve image aesthetics problems, which have demonstrated promising results. Recently, multi-modal large language models (MLLMs) have shown superior aesthetic perception and robustness in the fields of image aesthetics, greatly surpassing lightweight models due to their vast knowledge base and strong reasoning and memory capabilities.~\citep{huang2024aesexpert,huang2024aesbench,wu2024multimodal}.

\subsection{LLM Preferences Data}
Preference learning is an optimization method for LLMs designed to enhance their ability to generate outputs that better align with human preferences~\citep{furnkranz2010preference,ppo,dpo,ouyang2022training}. Increasing attention has also been drawn to the importance of data used during the preference learning phase. Some studies focus on constructing domain-specific datasets for preference learning, e.g., summarization~\citep{stiennon2020learning,wu2021recursively} and question answering~\citep{nakano2021webgpt}.~\citet{cui2024ultrafeedback} highlights the scarcity of large-scale, general-purpose preference datasets and propose UltraFeedback to addresses this gap by collecting over 1 million preference feedback samples using GPT-4~\citep{ChatGPT}. \citet{lee2023rlaif} also pointed out that utilizing AI-generated preference feedback is an effective and cost-efficient method for expanding preference datasets.
While the aforementioned work provides preference datasets for specific domains as well as general-purpose tasks, none of them have addressed the critical area of text aesthetics in LLMs, which motivated us to design corresponding data construction pipeline and related dataset like \ourdata{} to support future research in text aesthetics.

\section{Textual Aesthetics}
\subsection{Overview}
Textual aesthetics, which encompass the aesthetic attributes of a text at both the content and visual levels, can be dissected into four fundamental aspects. \textbf{Clarity} (readability) pertains to the ease with which a text can be read and comprehended, necessitating optimal sentence length and grammatical complexity \citep{dubay2004principles}. \textbf{Layout} (visual organization) involves the systematic arrangement of text elements, such as headings and subheadings, to guide the reader effectively. \textbf{Uniformity} (consistency) demands a consistent style and formatting throughout the text to enhance readability and facilitate a smoother reading experience. \textbf{Coherence} (overall structure) ensures that paragraphs are well-organized and logically connected, facilitating easier comprehension of the content \citep{van2014connectives}.


\subsection{Aesthetics Polishing}
\label{sec:polish}

Human preference data is critical for aligning large language models and improving their performance across various dimensions, such as helpfulness~\citep{askell2021general,kreutzer2018can,stiennon2020learning}, harmlessness~\citep{bai2022training,glaese2022improving}, and honesty~\citep{ouyang2022training}. Consequently, we believe that a textual aesthetic preference dataset will also be beneficial for research on the alignment of LLMs. However, current literature reveals a conspicuous absence of research specifically addressing the textual aesthetics of LLMs, as well as a lack of corresponding textual aesthetic preference data. To address this gap, we have developed a method for textual aesthetic polishing to construct a dataset that optimizes the aesthetic preferences of LLMs.

\begin{wrapfigure}{r}{0.6\textwidth}
    \vspace{-10pt}
    \includegraphics[width=0.6\textwidth]{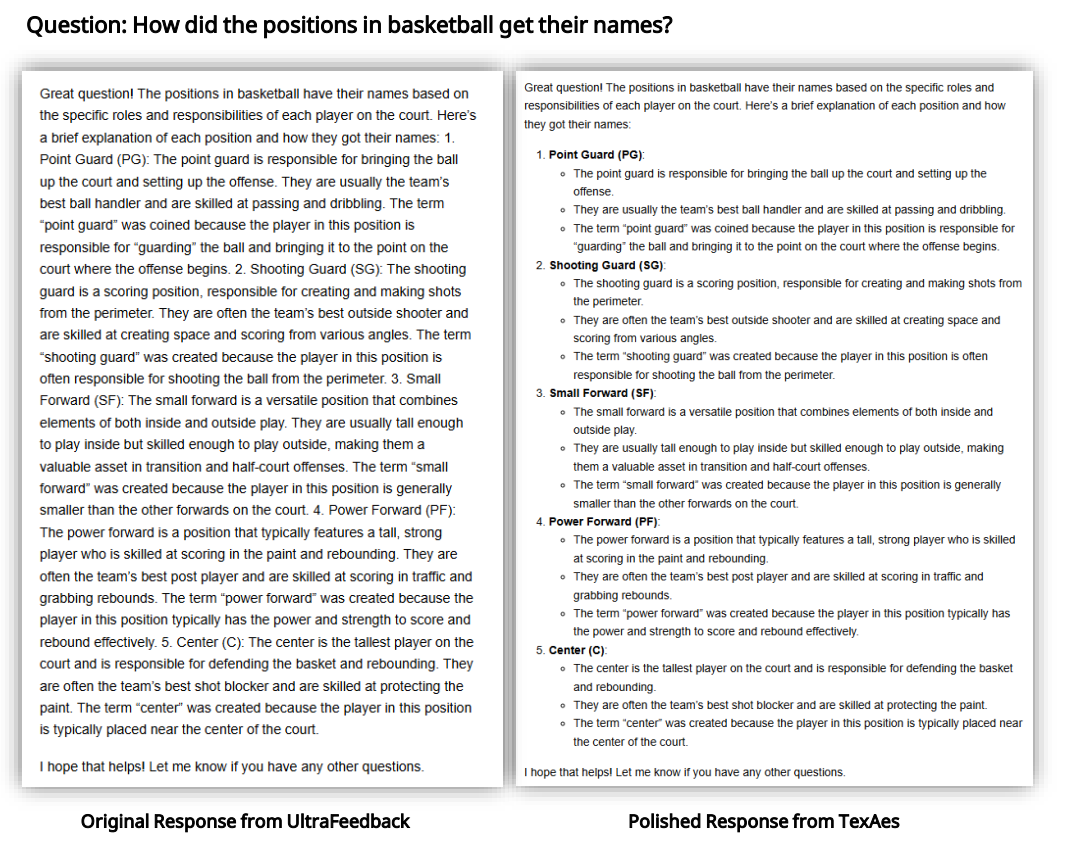} 
    \caption{Comparison of responses between the UltraFeedback and \ourdata{} datasets.}
    \label{fig:ta_case} 
    \vspace{-8mm}
\end{wrapfigure}

Given that the goal of polishing is to enhance textual aesthetics, we can build our textual aesthetic preference dataset based on an available preference dataset such as UltraFeedback~\citep{cui2024ultrafeedback}. UltraFeedback is a comprehensive dataset with responses evaluated by GPT-4 based on criteria such as instruction-following, honesty, and helpfulness. Since the selected data exhibits higher scores in these areas, thereby aligning more closely with human preferences, we can utilize these chosen responses as our candidates to build our textual aesthetic preference dataset.

To effectively achieve our objectives, we designed a chain of thought~\citep{wei2023chainofthoughtpromptingelicitsreasoning} methodology by using GPT-4o to polish our original responses, the following steps were taken:
\begin{enumerate}[leftmargin=1.5em,itemsep=0pt,parsep=0.2em,topsep=0.0em,partopsep=0.0em]
    \item \textbf{Semantic Analysis}: GPT-4o initially analyzed the textual semantics of the provided instructions and selected responses.
    \item \textbf{Aesthetic Evaluation}: Based on textual aesthetic factors such as paragraph structure, indentation, headings, and subheadings, GPT-4o conducted a detailed textual aesthetic analysis.
    \item \textbf{Binary Classification}: GPT-4o then performed a binary classification to determine whether the response required modification to improve readability and comprehension.
    \item \textbf{Revision Process}: For responses that required modification, GPT-4o generated a revised version that preserved the original style and format while enhancing readability and comprehensibility. The prompts requiring aesthetic modifications are documented in the appendix~\ref{app-prompt-polish}, which includes the text before and after modification, demonstrating the enhanced readability and comprehension of the revised text.
\end{enumerate}
Finally, we compiled the textual aesthetic preference dataset \(\mathcal{D} = \left\{ \left( x^{(i)}, y_{t}^{(i)}, y_{w}^{(i)}, y_{l}^{(i)} \right) \right\}_{i=1}^{N} \), where \( y_{t}^{(i)} \) represents the revised textual aesthetic data, and \( y_{w}^{(i)} \) and \( y_{l}^{(i)} \) represent the originally chosen and rejected data in UltraFeedback, respectively.

We observed that some polished responses became overly verbose and less natural or human-like. We hypothesize that this is because the original responses in UltraFeedback are already of high quality, making the task of polishing more challenging than expected. To address this issue, we implemented a length constraint for the polishing process. Future work will focus on further improving the textual aesthetic polishing method.

\subsection{Textual Aesthetics Scoring}
\label{sec:eval}

To validate the aesthetic quality of texts generated by large language models and to assess the effectiveness of our aesthetic preference dataset, a robust method for evaluating text aesthetics is indispensable. Previous studies, such as AlpacaEval \citep{alpaca_eval,dubois2024length}, MT-Bench \citep{zheng2023judging}, and Arena-Hard \citep{li2024crowdsourced}, suggest that using LLMs as evaluators can effectively approximate human preferences. Consequently, we employ the "LLM as a judge" framework to approximate human preferences for text aesthetics. We evaluate the aesthetic quality of texts generated by LLMs using two methods: text-based and image-based text aesthetic scoring.

\textbf{Text-Based Text Aesthetic Scoring.} We randomly selected 500 prompts from Arena-Hard~\citep{li2024crowdsourced} as our evaluation dataset. Following practices from Arena-Hard and MT-Bench~\citep{zheng2023judging}, we implemented a pairwise comparison method, comparing the performance of model $\pi_i$ on prompt $p$ with a robust baseline model (GPT-4-0314) to derive aesthetic preference scores. Judges assessed aesthetic preferences on a Likert scale \citep{likert1932technique} (1 = prefers $\pi_i(p)$ much less than $\pi_{\text{base}}(p)$, 5 = prefers $\pi_i(p)$ much more than $\pi_{\text{base}}(p)$). This methodology ensures that models are penalized more heavily for substantial losses than for minor ones, effectively differentiating between models. Using the chain-of-thought approach, judges evaluated text aesthetics based on four dimensions: readability, visual organization, consistency, and overall structure. To mitigate position bias, we employed a two-game setup by swapping model positions for each query. Following the practices of Chatbot Arena, we adopted the Bradley-Terry \citep{bradley1952rank} model to generate final scores. We aggregated all pairwise comparisons with the baseline model and employed bootstrapping to derive a bootstrapped confidence interval for all models’ win rates against the baseline, producing an ordered ranking of all models based on their win rates. The judge prompts are provided in Appendix~\ref{app-prompt-texteval}.

\textbf{Image-Based Text Aesthetic Scoring.} Our conceptualization of text aesthetics encompasses not only textual readability and comprehensibility but also visual appeal. Given GPT-4o's exceptional multimodal capabilities, we utilized GPT-4o to evaluate text aesthetics from a visual perspective as well. In our experiments, we rendered the LLM-generated texts as HTML with consistent CSS styles, converted them into images of identical size, and then had GPT-4o evaluate these images based on the same criteria used for textual evaluation. Specific prompts are provided in Appendix~\ref{app-prompt-imageeval}.

\section{Textual Aesthetics-Powered Training}
\subsection{Direct Preference Optimization Training}

Reinforcement Learning with Human Feedback (RLHF)\citep{christiano2017deep} has emerged as a pivotal technique in aligning LLMs~\citep{bai2022training, ouyang2022training, stiennon2020learning}. Early implementations of RLHF primarily relied on reinforcement learning and alternative approaches~\citep{snell2022offline,touvron2023llama,gulcehre2023reinforced}. \citet{rafailov2024directpreferenceoptimizationlanguage} proposed a RL-free closedform counterpart known as Direct Preference Optimization (DPO) which has shown impressive performances~\citep{ivison2023camels,jiang2023mistral,tunstall2023zephyr}.

The naive DPO uses a pair of preference data, which includes a chosen response and a rejected response for each prompt, based on the Bradley-Terry \citep{bradley1952rank} model for optimization. The loss function for DPO is defined as follows:

\vspace{-1em} 
\begin{equation}
\mathcal{L}_{\text{DPO}}(\pi_\theta; \pi_{\text{ref}}) = - \mathbb{E}_{(x, y_w, y_l) \sim \mathcal{D}} \left[ \log \sigma \left( \beta \log \frac{\pi_\theta(y_w \mid x)}{\pi_{\text{ref}}(y_w \mid x)} - \beta \log \frac{\pi_\theta(y_l \mid x)}{\pi_{\text{ref}}(y_l \mid x)} \right) \right],
\label{eq:dpo_loss}
\end{equation}
\vspace{-1em} 

where \(\pi_\theta\) denotes the policy being optimized, \(\pi_{\text{ref}}\) represents the reference policy, \(x\) is the input prompt, \(y_w\) is the chosen (winning) response, \(y_l\) is the rejected (losing) response, \(\mathcal{D}\) is the dataset of prompts and responses, \(\sigma\) is the sigmoid function, and \(\beta\) is a scaling parameter.  By directly integrating preference data into the optimization process, DPO ensures that the generated text aligns closely with human judgments.

\subsection{\ourmethodfullname{} Training}
\label{sec:TAPO_define}
For each prompt in our \ourdata{} dataset, there are three responses: \( y_t \), \( y_w \), and \( y_l \). The response \( y_t \) has the same semantic content as \( y_w \) but is superior in terms of textual aesthetics. The response \( y_w \), in turn, is more aligned with human preferences for chatbots in terms of instruction-following, truthfulness, honesty, and helpfulness compared to \( y_l \). The response \( y_l \) is the least preferred response in terms of both textual aesthetics and human preferences. The goal of our training is to learn a model that can generate responses that are both aesthetically pleasing and preferred by humans. To achieve this, we designed a textual aesthetics preference optimization (\ourmethod{}) approach that jointly optimizes for both textual aesthetics and human preferences.

To simultaneously utilize all three preference data types in the \ourdata{} dataset for optimization, we adopt the Plackett-Luce~\citep{luce1959individual,plackett1975analysis} model as the underlying preference model. \citet{rafailov2024directpreferenceoptimizationlanguage} showed that \(\beta \log \frac{\pi_\theta(y \mid x)}{\pi_{\text{ref}}(y \mid x)}\) can be treated as “implicit reward” which is assumed to represent the preference for the model generate \(y\) given the prompt \(x\), the goal of DPO is to align the “implicit reward” towards human preference data directly. We denote each reward function \(\beta \log \frac{\pi_\theta(y_k \mid x)}{\pi_{\text{ref}}(y_k \mid x)}\) as \(r_{\theta}(x, y_k)\) (where \(k \in \{t, w, l\}\)), representing the preferences for the model generating \(y_t\), \(y_w\) \(y_l\) given the input \(x\). \(\pi_{\theta}\) and \(\pi_{\text{ref}}\) are the policy model and reference model respectively and \(\beta\) is a hyper-parameter to control the KL divergence between \(\pi_\theta\) and \(\pi_{\text{ref}}\). The training objective of TAPO is

\vspace{-1em} 
\begin{equation}  
    \begin{split}  
    &\mathcal{L}_{\text{\ourmethod{}}}(\pi_{\theta}; \pi_{\text{ref}}) = \\
    &-\mathbb{E}_{(x, y_t, y_w, y_l) \sim \mathcal{D}} \left[ \log \left( \frac{\exp(r_{\theta}(x, y_t))}{\sum_{i \in \{t, w, l\}} \exp(r_{\theta}(x, y_i))} \cdot \frac{\exp(r_{\theta}(x, y_w))}{\sum_{i \in \{w, l\}} \exp(r_{\theta}(x, y_i))} \right) \right]
    \end{split}  
\end{equation}
\vspace{-1em} 

where $\mathcal{D}$ is the dataset, and $\beta$ is the temperature parameter. 

Using the properties of logarithmic functions, the loss function can be decomposed into two parts: \(\mathcal{L}_\text{TA}\) and \(\mathcal{L}_\text{DPO}\):

\vspace{-2em} 
\begin{equation}
    \mathcal{L}_\text{TA} = -\log \left(  \frac{\exp(r_{\theta}(x, y_t))}{\sum_{i \in \{t, w, l\}} \exp(r_{\theta}(x, y_i))}\right), \quad\mathcal{L}_\text{DPO} = -\log \left( \frac{\exp(r_{\theta}(x, y_w))}{\sum_{i \in \{w, l\}} \exp(r_{\theta}(x, y_i))} \right).
\label{eq:loss_two_part}
\end{equation}
\vspace{-1em} 

It can be observed that \(\mathcal{L}_\text{DPO}\) is identical to the loss used in Bradley-Terry model-based preference optimization with \( y_w \) and \( y_l \), as demonstrated in the proof provided in Appendix~\ref{appendix:proof_equivalence}. On the other hand, \(\mathcal{L}_\text{TA}\) represents the log probability of \(r_{\theta}(x, y_t)\) being ranked first among \(r_{\theta}(x, y_t)\), \(r_{\theta}(x, y_w)\), and \(r_{\theta}(x, y_l)\). \(\mathcal{L}_\text{DPO}\) primarily optimizes the model's preference for honest, helpful, and truthful data, whereas \(\mathcal{L}_\text{TA}\) optimizes both the correctness of the answers and textual aesthetics. To ensure the generated answers are not only accurate but also aesthetically pleasing, we assign different weights to the losses to adjust the preference optimization direction. The modified loss function is as follows:

\vspace{-1em} 
\begin{equation}
\mathcal{L}_{\text{\ourmethod{}}}(\pi_{\theta}, \pi_{\text{ref}}) = -\mathbb{E}_{(x, y_t, y_w, y_l) \sim \mathcal{D}} \left[ w_\text{TA} \cdot \mathcal{L}_\text{TA} + w_\text{DPO} \cdot \mathcal{L}_\text{DPO} \right].
\end{equation}
\vspace{-1em} 

\section{Data and Experiment Settings}

\subsection{Textual Aesthetics Dataset}

\begin{wraptable}{r}{0.4\textwidth}
\centering
\resizebox{\linewidth}{!}{
\renewcommand\tabcolsep{6.0pt}
\begin{tabular}{lcc}
\toprule
{\textbf{Dataset}} & \textbf{\#Prompts} & \makecell{\textbf{Response} \\ \textbf{Length}} \\
\midrule
\textsc{UltraFeedback} & 61,135  & 297\\
\ourdata{} & 50,390  &  293\\
 \bottomrule
\end{tabular}}
\caption{Statistics of \ourdata{} datasets.}
\label{tab:data}
\vspace{-5mm}
\end{wraptable}
As introduced in Section~\ref{sec:polish}, we constructed our textual aesthetic dataset based on a filtered version of UltraFeedback\footnote{\url{https://huggingface.co/datasets/HuggingFaceH4/ultrafeedback_binarized}}~\citep{cui2024ultrafeedback,ivison2023camels,tunstall2023zephyr} dataset, which comprises 61,135 completions evaluated by GPT-4, including both accepted and rejected entries. In our experiment, we utilized GPT-4o to perform aesthetic polishing on the UltraFeedback dataset.
After the aesthetic polishing process, we found that 5,858 entries were already aesthetically satisfactory and required no further modification. We then analyzed the length of the filtered texts and discovered that a minor subset exhibited excessive verbosity and lacked a natural, human-like quality. To address this, we excluded outliers in the distribution of length differences before and after aesthetic polishing, retaining only data within the 90\% confidence interval. We present the statistics of \ourdata{} in Table~\ref{tab:data}. We present the length distribution in appendix~\ref{app:datastat} and length constraint filter experiment in appendix~\ref{app:data_length}.

\subsection{Experiment Settings}
\label{exp:eval}
In this study, we evaluate the performance of models from two perspectives: textual aesthetics and general response capabilities.
For textual aesthetics, we compare the models using both text-based and image-based text aesthetic scoring methods, as described in Section~\ref{sec:eval}. We report the win rate (WR) in text aesthetics at both the text and image levels relative to the baseline model (GPT-4-0314). In addition to automatic evaluation, we conduct a human evaluation to further validate the models' performance. We randomly sample fifty entries from the Anera-Hard~\cite{} dataset and ask human annotators to rate the aesthetics of these entries. 

To evaluate the changes in the model's general capabilities following the alignment of textual aesthetics preferences, including its ability to follow instructions and respond to complex prompts across diverse domains, we utilize three well-established auto-evaluation instruction-following benchmarks based on GPT-4-as-a-Judge: AlpacaEval 2.0~\citep{dubois2024length}, Arena-Hard~\citep{li2024crowdsourced} and MT-Bench~\citep{zheng2023judging}. For both the supervised tine-tuning and \ourmethod{} stages, we employ a low-rank adaptation~\citep{hu2021loralowrankadaptationlarge} adapter instead of fine-tuning the entire model. Detailed training parameters are provided in the appendix~\ref{app:param}.
\vspace{-2mm}

\section{Experiment Results}
\subsection{Main Results}
The comparative analysis of our models trained with \ourmethod{} on \ourdata{} against open-source models is shown in Table~\ref{tbl:main}. Our \ourlrgmodel{} model surpasses all open-source counterparts in both text-based and image-based text aesthetic metrics, with an 18.88\% improvement in text-based scores and a 27.85\% enhancement in image-based scores over the best-performing LLaMA-3.1-70B-Instruct model. 

For general response benchmarks, the LLaMA-3.1-8B-Instruct and LLaMA-3.1-70B-Instruct models, after \ourmethod{} training, show improvements on AlpacaEval 2.0 and MT-Bench, though with a slight decline on Arena-Hard. AlpacaEval 2.0 focuses on chat scenarios, MT-Bench on multi-turn conversations, and Arena-Hard on more complex queries. The gains in AlpacaEval 2.0 and MT-Bench suggest that enhanced text aesthetics contribute to better conversational abilities, aligning with our goal of improving answer clarity, layout, uniformity, and coherence. This underscores the quality of \ourdata{} and the effectiveness of \ourmethod{} in boosting both text aesthetics and overall model performance.

\begin{table}[h]
    \centering
    \caption{Performance comparison between \ourmethod{} models and open-source models across various benchmarks.``TA Text'' and ``TA Image'' denote text-based and image-based textual aesthetic metrics, respectively. Metrics include: win rates (WR) against GPT-4-Turbo for TA Text and TA Image, WR against GPT-4-0314 for Arena-Hard, length-controlled (LC) win rate against GPT-4-Turbo in AlpacaEval 2.0, and average scores for MT-Bench. All evaluations are conducted using GPT-4 as the judge.}
    \resizebox{\linewidth}{!}{
    \renewcommand\tabcolsep{6.0pt}
    \begin{tabular}{lcccccc}
    \toprule
        \textbf{Model} & \textbf{Size} & \makecell[c]{\textbf{TA Text} \\ \textbf{WR(\%)}} & \makecell[c]{\textbf{TA Image} \\ \textbf{WR(\%)}} & \makecell[c]{\textbf{AlpacaEval 2.0} \\ \textbf{LC WR(\%)}} & \makecell[c]{\textbf{Arena-Hard} \\ \textbf{WR(\%)}} & \makecell[c]{\textbf{MT-Bench} \\ \textbf{Avg. Score}} \\ 
        \midrule
        Qwen2-7B-Instruct~\citep{qwen2} & 7B & 24.63 & 39.40 & 33.43 & 27.69 & 7.48 \\ 
        Yi-1.5-9B-Chat~\citep{ai2024yi} & 9B & 35.52 & 55.03 & 34.74 & \textbf{38.89} & 7.38\\ 
        LLaMA-3.1-8B-Instruct~\citep{dubey2024llama3herdmodels} & 8B & 33.42 & 47.94  & 41.34 & 37.10 & 7.42 \\ 
        \ourmodel{} & 8B & \textbf{50.85} & \textbf{71.91} & \textbf{49.84} & 33.89 & \textbf{7.72} \\ \midrule
        Tulu-2-dpo-70B~\citep{ivison2023camels} & 70B & 9.43 & 27.79  & 31.01 & 16.37 & 6.89 \\ 
        Qwen2-72B-Instruct~\citep{qwen2} & 72B & 22.05 & 30.68  & 40.61 & 42.48 & 8.22 \\ 
        LLaMA-3.1-70B-Instruct~\citep{dubey2024llama3herdmodels} & 70B & 53.18 & 57.34 & 45.03 &\textbf{ 67.22} & 8.16 \\ 
        \ourlrgmodel{} & 70B & \textbf{63.22} & \textbf{73.31} & \textbf{51.26} & 63.42 & \textbf{8.30}\\ \bottomrule
    \end{tabular}
    }
    \label{tbl:main}
    \vspace{-4mm}
\end{table}

\begin{figure*}[t] 
\centering
\includegraphics[width=0.49\textwidth]{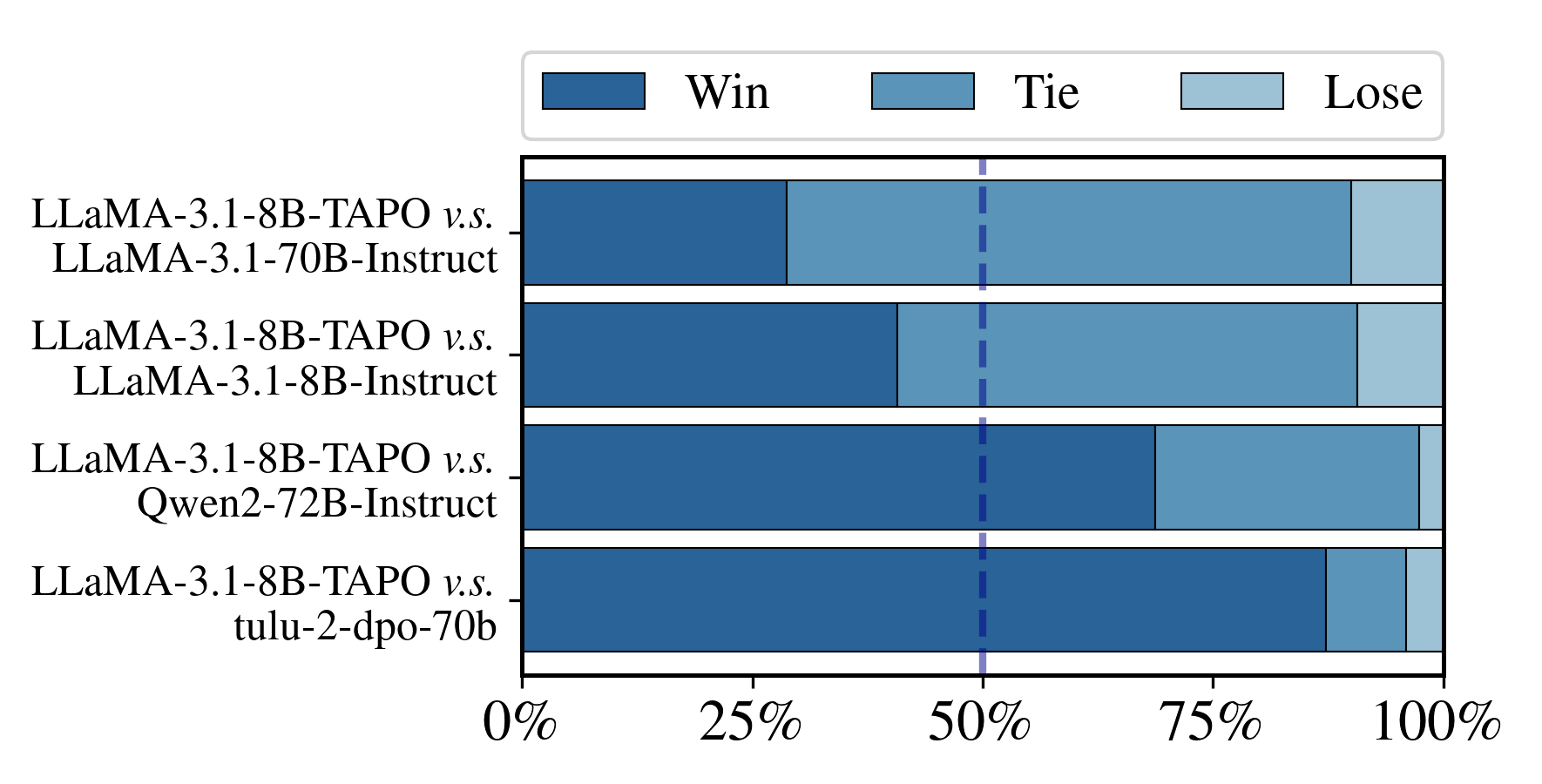}
\includegraphics[width=0.49\textwidth]{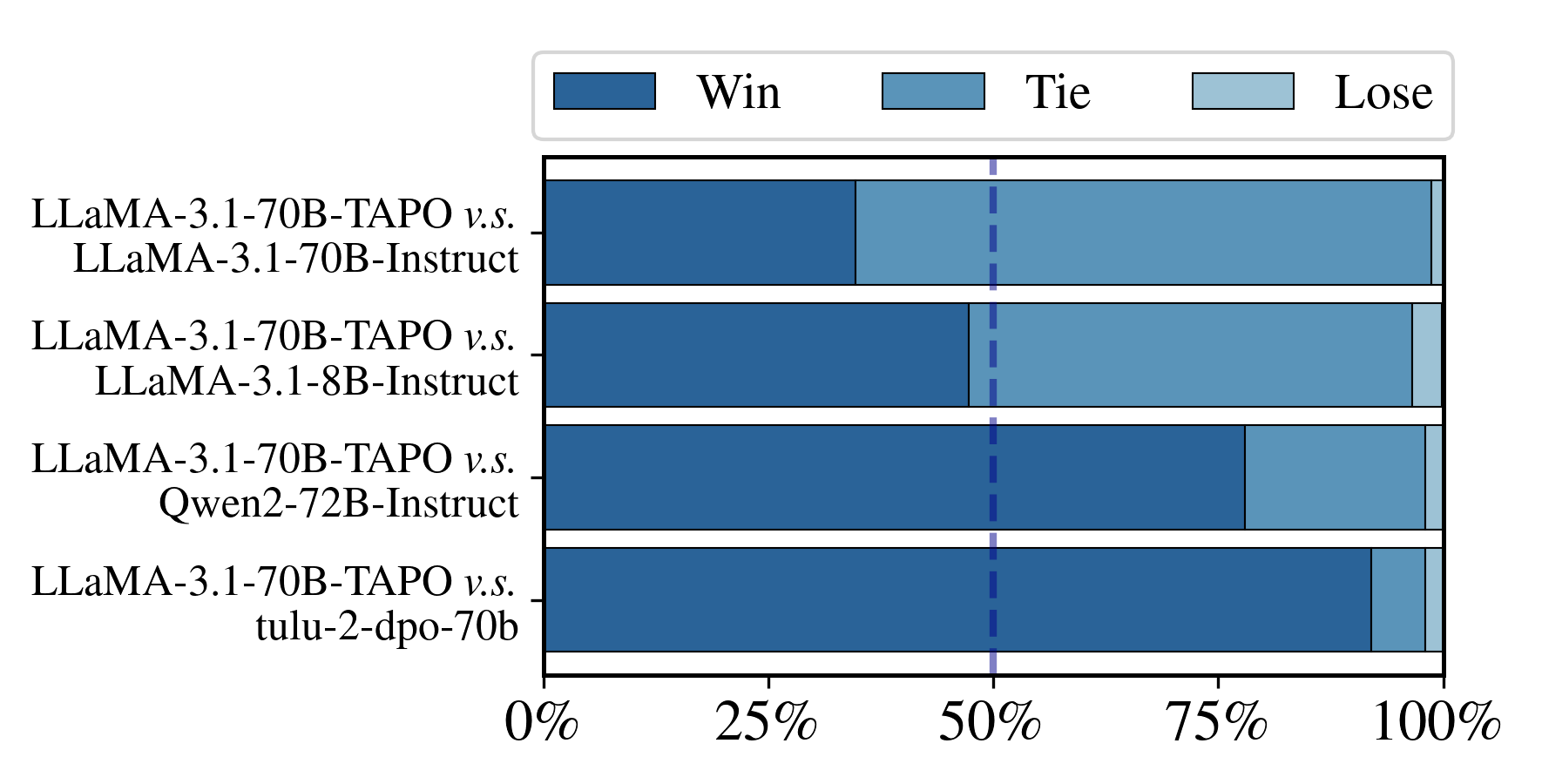}
\\
\vspace{-1mm}
\makebox[0.49\textwidth]{\small \ourmodel{} \emph{v.s.} Others}
\makebox[0.49\textwidth]{\small \ourlrgmodel{} \emph{v.s.} Others}
\caption{Win rates of models fine-tuned by \ourmethod{} compared to other SOTA open-source models by human judgements in textual aesthetics level. Human judgments are majority votes from three annotators.} 
\label{fig:Win rate VideoRM}
\vspace{-4mm}
\end{figure*}

The results of the human evaluation, shown in Figure~\ref{fig:Win rate VideoRM}, show that our \ourlrgmodel{} model is rated significantly higher in text aesthetics than the best-performing open-source model. These results confirm that our model is more visually appealing and coherent, consistent with our quantitative analysis, further validating the efficacy of \ourmethod{} in enhancing text aesthetics and overall performance.

\subsection{Impact of Loss Weight}

To determine the influence of the weight ratio between \(\mathcal{L}_\text{TA}\) and \(\mathcal{L}_\text{DPO}\) in \ourmethod{} on the aesthetics of the text of the model and the overall performance, we performed a series of methodical experiments. Specifically, we experimented with two settings: 1. First, we used the Tulu-v2 dataset~\citep{ivison2023camels} to fine-tune the \textit{LLaMA-3.1-8B-base} model in a supervised manner, followed by further optimization using \ourmethod{}; 2. Second, we directly applied \ourmethod{} to the \textit{LLaMA-3.1-8B-instruct} model.
We set the weight ratios of \(\mathcal{L}_\text{TA}\) to \(\mathcal{L}_\text{DPO}\) at 2:1, 1:1, 1:2 and 1:5, respectively, to train the models. We then evaluated the models' text-based and image-based text aesthetic scores, as well as their performance on Arena-Hard.

\begin{figure}[h]
    \centering
    \begin{subfigure}{0.38\textwidth}
        \centering
        \includegraphics[width=\textwidth]{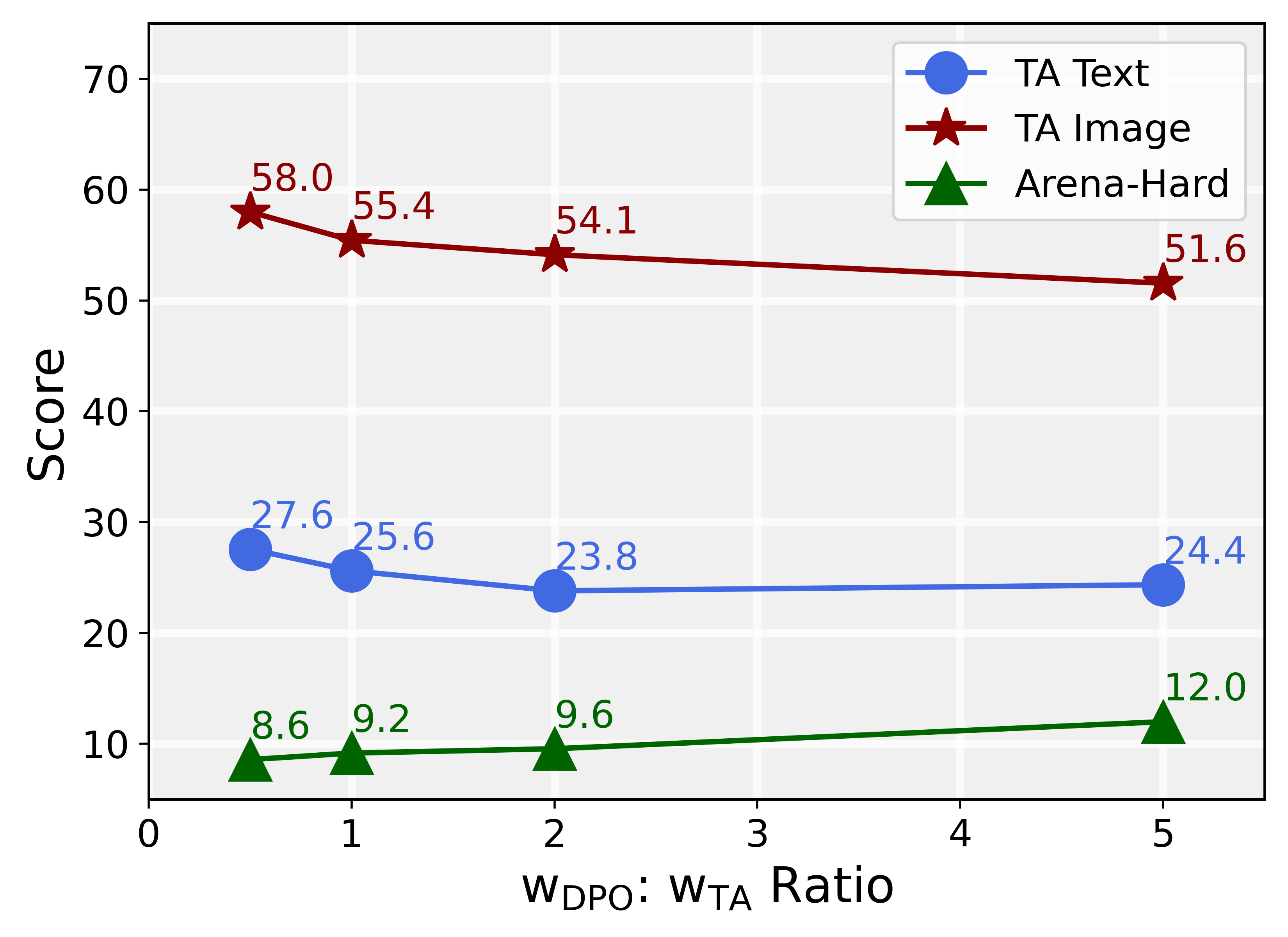} 
        \caption{LLaMA-3.1-8B-Base}
        \label{fig:ta_scores_base}
    \end{subfigure}
    \hspace{2mm}
    \begin{subfigure}{0.38\textwidth}
        \centering
        \includegraphics[width=\textwidth]{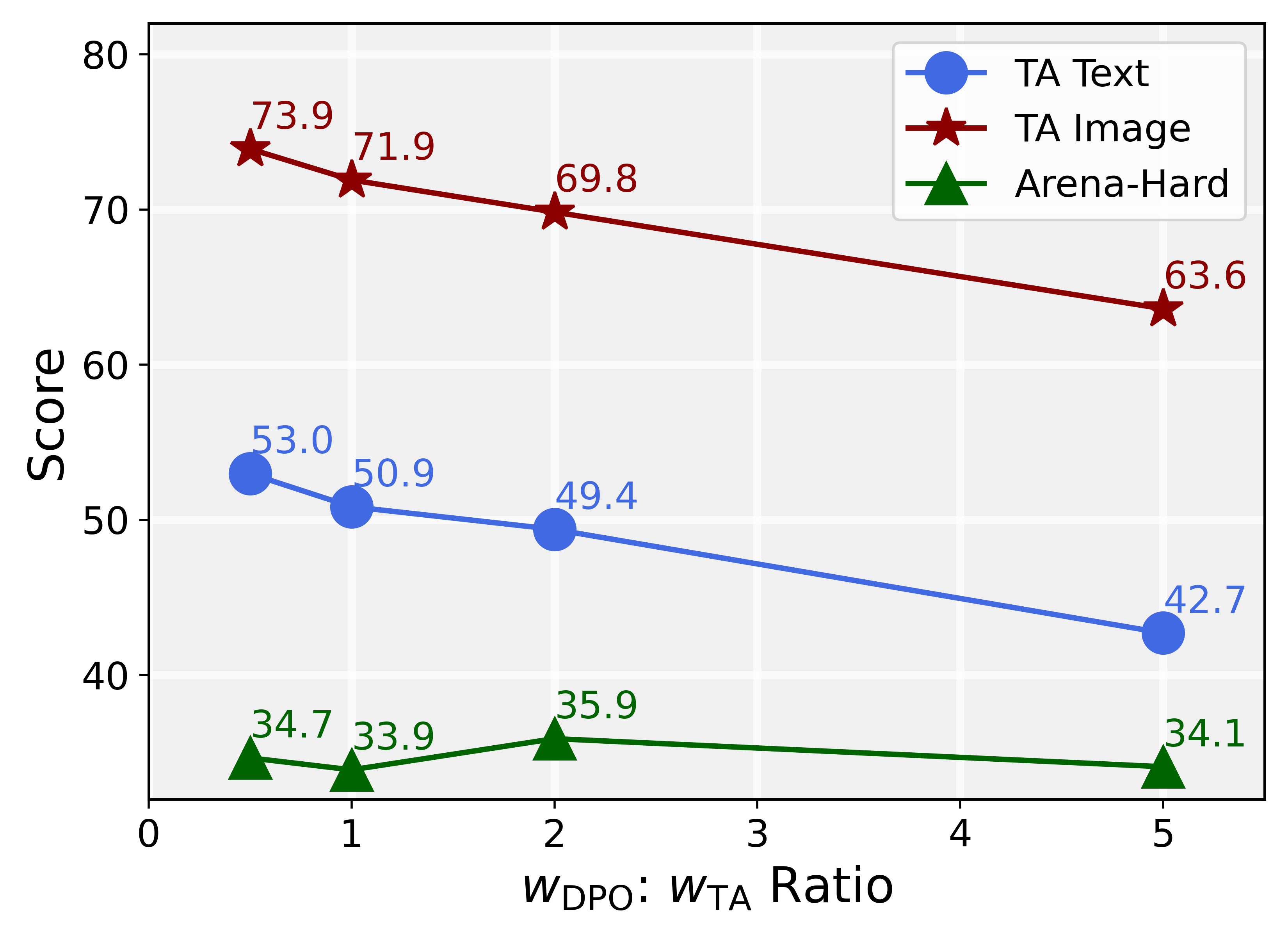} 
        \caption{LLaMA-3.1-8B-Instruct}
        \label{fig:ta_scores_instruct}
    \end{subfigure}
    \caption{Performance Across Various Weight Ratios}
    \label{fig:weight_ratio}
    \vspace{-3mm}
\end{figure}

Figures~\ref{fig:ta_scores_base} and \ref{fig:ta_scores_instruct} illustrate the performance variations of \ourmethod{} across different weight ratios. For the \textit{LLaMA-3.1-8B-base} model, increasing the proportion of \(\mathcal{L}_\text{DPO}\) consistently improves the Arena-Hard score but decreases both text-based and image-based text aesthetic scores. This indicates that a higher proportion of \(\mathcal{L}_\text{DPO}\) improves optimization toward human preference at the expense of aesthetic preference.
For the \textit{LLaMA-3.1-8B-instruct} model, which is already aligned with human preferences, further increasing \(\mathcal{L}_\text{DPO}\) yields limited improvements in instruct-following capability and significantly decreases textual aesthetic preference.

\subsection{Two-Stage Training}

To validate the efficacy of incorporating three types of preference data in \ourmethod{}, we conducted a two-stage DPO training ablation experiment. Initially, human preferences were aligned using the \( y_w \) and \( y_l \) data sets, denoted as \(\text{DPO}(y_w, y_l)\). Subsequently, text aesthetic preference alignment was conducted using two methods: \(\text{DPO}(y_t, y_w)\) and \(\text{DPO}(y_t, y_l)\). These experiments were performed on the \textit{LLaMA-3.1-Base} and \textit{LLaMA3.1-Instruct} models, with results presented in Table \ref{tab:two_stage}.

Comparing the final models from the two-stage training with our model trained in \ourmethod{} method, we found that, except for the image-based text aesthetic metric, where our model was slightly inferior, it significantly outperformed the two-stage models on text-based aesthetic metrics, AlpacaEval 2.0, Arena-Hard, MT-Bench, and MMLU\citep{hendrycks2020measuring}. This suggests that \ourmethod{}, by leveraging three types of preference data, not only enhances text aesthetic scores but also improves general capabilities.

\begin{table}[t]
    \centering
    \caption{Comparison of two-stage DPO training and \ourmethod{} training. DPO($y_w, y_l$) + DPO($y_t, y_l$) denotes two-stage training where the first stage is DPO($y_w, y_l$) and the second stage is DPO($y_t, y_l$). The LLaMA-3.1-8B-Base is fine-tuned using the Tulu-v2 dataset.}
    \resizebox{0.7\linewidth}{!}{
    \renewcommand\tabcolsep{4.0pt}
    \begin{tabular}{lcccccc}
    \toprule
         \makecell{\textbf{Training Settings}} & \makecell[c]{\makecell{\textbf{TA} \textbf{Text}} \\ \textbf{WR(\%)}} & \makecell[c]{\makecell{\textbf{TA} \textbf{Image}} \\ \textbf{WR(\%)}} & \makecell[c]{\textbf{AlpacaEval 2.0} \\ \textbf{LC WR(\%)}} & \makecell[c]{\textbf{Arena-Hard} \\ \textbf{WR(\%)}} & \makecell[c]{\textbf{MT-Bench} \\ \textbf{Avg. Score}} & \makecell[c]{\textbf{MMLU} \\ \textbf{5-shot}} \\
        \midrule
        \multicolumn{7}{c}{\textbf{LLaMA-3.1-8B-Base}} \\
        \midrule
         DPO($y_w, y_l$) + DPO($y_t, y_l$) & 25.45 & \textbf{60.53} & 23.77 & 7.72 & 5.98 & 63.52 \\
         DPO($y_w, y_l$) + DPO($y_t, y_w$) & 14.03 & 48.31 & 14.66 & 5.35 & 5.50 & 62.80 \\ 
         \ourmethod{}($y_t, y_w, y_l$) & \textbf{25.61} & 55.43 & \textbf{26.05} & \textbf{9.16} & \textbf{6.05} & \textbf{64.48} \\   
        \midrule \multicolumn{7}{c}{\textbf{LLaMA-3.1-8B-Instruct}} \\ \midrule
         DPO($y_w, y_l$) + DPO($y_t, y_l$) & 50.26 & 71.33 & 46.47 & 31.08 & \textbf{7.75} & 68.41 \\ 
         DPO($y_w, y_l$) + DPO($y_t, y_w$) & 50.76 & \textbf{75.69} & 44.91 & 29.41 & 7.39 & 67.89 \\ 
         \ourmethod{}($y_t, y_w, y_l$) & \textbf{50.85} & 71.91 & \textbf{49.84} & \textbf{33.89} & 7.72 & \textbf{68.80} \\ \bottomrule
    \end{tabular}
    }
    \label{tab:two_stage}
    \vspace{-4mm}
\end{table}

\subsection{\ourdata{} vs. UltraFeedback}
To validate the effectiveness of the \ourdata{} data set, we performed a comparative analysis of models trained using \ourdata{} against those trained with UltraFeedback data. We applied the Direct Preference Optimization (DPO) method to align human preferences with the \( y_w \) and \( y_l \) pairs from UltraFeedback and the \( y_t \) and \( y_l \) pairs from \ourdata{}. The experiments were conducted on both the \textit{LLaMA-3.1-Base} and \textit{LLaMA3.1-Instruct} models.

\begin{table}[h]
    \centering
    \caption{Comparative analysis of \ourdata{} and UltraFeedback with DPO Training. The baseline represents the performance of LLaMA-3.1-8B-Base which is fine-tuned using the Tulu-v2 dataset and \ourbasemodel{}.}
    \resizebox{0.7\linewidth}{!}{
    \renewcommand\tabcolsep{4.0pt}
    \begin{tabular}{lcccccc}
    \toprule
        \makecell{\textbf{Dataset}} & \makecell[c]{\makecell{\textbf{TA} \textbf{Text}} \\ \textbf{WR(\%)}} & \makecell[c]{\makecell{\textbf{TA} \textbf{Image}} \\ \textbf{WR(\%)}} & \makecell[c]{\textbf{AlpacaEval 2.0} \\ \textbf{LC WR(\%)}} & \makecell[c]{\textbf{Arena-Hard} \\ \textbf{WR(\%)}} & \makecell[c]{\textbf{MT-Bench} \\ \textbf{Avg. Score}} & \makecell[c]{\textbf{MMLU} \\ \textbf{5-shot}} \\
        \midrule
        \multicolumn{7}{c}{\textbf{LLaMA-3.1-8B-Base}}  \\
        \midrule
        Baseline & 1.17 & 8.60 & 5.24 & 4.10 & 5.60 & 64.07 \\ 
        UltraFeedback & 2.56 & 8.17 & 9.29 & 7.06 & \textbf{5.92} & \textbf{65.02} \\ 
        \ourdata{} & \textbf{25.79} & \textbf{60.64} & \textbf{24.06} & \textbf{9.04} & 5.78 & 63.17 \\ \midrule
        \multicolumn{7}{c}{\textbf{LLaMA-3.1-8B-Instruct}}  \\ \midrule
        Baseline & 33.42 & 47.94 & 41.34 & \textbf{37.10} & 7.42 & 68.80 \\ 
        UltraFeedback & 30.92 & 48.57 & 44.19 & 34.74 & \textbf{7.76} & \textbf{68.90} \\ 
        \ourdata{} & \textbf{49.07} & \textbf{68.63} & \textbf{45.82} & 29.87 & 7.55 & 68.52 \\ \bottomrule
    \end{tabular}
    }
    \label{tab:ta_vs_uf}  
\end{table}

The results, shown in Table \ref{tab:ta_vs_uf}, indicate that for the \textit{LLaMA-3.1-Base} model, UltraFeedback improved performance in AlpacaEval 2.0, Arena-Hard, MT-Bench and MMLU. For the \textit{LLaMA3.1-Instruct} model, there were performance improvements across most tasks, except for a slight decline in Arena-Hard. However, UltraFeedback did not improve performance in aesthetic evaluation tasks.
Models trained with \ourdata{} showed significant performance improvements over those trained with UltraFeedback in most tasks on the \textit{LLaMA-3.1-Base} model, with a minor decrease in MMLU. For the \textit{LLaMA3.1-Instruct} model, the one trained with \ourdata{} exhibited general capabilities comparable to those of the UltraFeedback-trained model while surpassing it in aesthetic tasks.
These experiments demonstrate that \ourdata{} not only optimizes the textual aesthetic performance of large language models but also aligns well with human preferences.

\vspace{-1em}
\subsection{Annotation Consistency}
We generated responses for 50 questions sampled from Arena-Hard using six models: \textit{\ourmodel{}}, \textit{\ourlrgmodel{}}, \textit{\ourbasemodel{}}~\citep{dubey2024llama3herdmodels}, \textit{\ourlrbasemodel{}}~\citep{dubey2024llama3herdmodels}, \textit{Qwen2-72B-Instruct}~\citep{qwen2}, and \textit{Tulu-2-dpo-70B}~\citep{ivison2023camels}. Subsequently, we employed three types of evaluators: text-based GPT-4o judge (TA Text), image-based GPT-4o judge (TA Image), and three human annotators, consisting of two graduate students and one professor. Each evaluator was tasked with comparing \textit{\ourmodel{}} and \textit{\ourlrgmodel{}} against other models in terms of the textual aesthetics of the generated answers (win/tie/lose), resulting in 400 annotated comparison pairs.

\begin{wraptable}{r}{0.45\textwidth}
    \centering  
    \vspace{-1em}
    \caption{Agreement between judges and human annotators on 400 samples from Arena-Hard. A-1, A-2, and A-3 are three human annotators. TA Text is the text-based GPT-4o judge, and TA Image is the image-based GPT-4o judge.} 
    \resizebox{\linewidth}{!}{  
    \renewcommand\tabcolsep{6.0pt}  
    \begin{tabular}{lcccc}  
    \toprule  
        \textbf{Judge} & \textbf{A-1} & \textbf{A-2} & \textbf{A-3} & \textbf{Average}  \\  
    \midrule  
        A-1 & - & 78.25\% & 77.50\% & 77.88\% \\ 
        A-2 & 78.25\% & - & 80.75\% & 79.50\% \\ 
        A-3 & 77.50\% & 80.75\% & - & 79.13\% \\ 
        TA Image & 60.75\% & 68.00\% & 65.75\% & 64.83\% \\ 
        TA Text & 69.00\% & 70.00\% & 67.00\% & 68.67\% \\
    \bottomrule  
    \end{tabular}  
    }  
    \label{tbl:judge_comparison} 
    \vspace{-1em}
\end{wraptable}

Table~\ref{tbl:judge_comparison} presents the agreement ratios, as utilized in MT-Bench~\citep{zheng2023judging}, among the TA Text scores, TA Image scores, and annotators, as well as annotators themselves. On average, the TA Text scores demonstrated a 68.67\% agreement rate with the human annotators, while the TA Image scores judges exhibited a 64.83\% agreement rate, which is lower than that of the human annotators. Notably, the agreement rates of both our image-based and text-based GPT-4o judges are comparable to those observed in previous human evaluations, which reported an average of 66\% agreement in MT-Bench~\citep{zheng2023judging} and 59.7\% in UltraFeedback~\citep{cui2024ultrafeedback}. These results suggest that our GPT-4o judges can serve as effective proxies for human preferences in assessing text aesthetics.
\vspace{-3mm}

\subsection{Criteria for Reject Sample Selection}
To effectively optimize textual aesthetics using preference optimization, it is essential to construct preference pairs consisting of chosen and rejected responses. For our purposes, we select \( y_t \) from \ourdata{} as the chosen response. As the rejected response, we use either the original chosen response \( y_w \) or the original rejected response \( y_l \) from the UltraFeedback dataset.
We conducted DPO experiments to compare the impact of \( y_w \) and \( y_l \) on the model's performance. The results are presented in Table \ref{tab:reject_chose}.

\begin{table}[h]
    \centering
    \caption{Evaluation of performance across different rejected samples.}
    \resizebox{0.7\linewidth}{!}{
    \renewcommand\tabcolsep{4.0pt}
    \begin{tabular}{lcccccc}
    \toprule
         \makecell{\textbf{Training Settings}} & \makecell[c]{\makecell{\textbf{TA} \textbf{Text}} \\ \textbf{WR(\%)}} & \makecell[c]{\makecell{\textbf{TA} \textbf{Image}} \\ \textbf{WR(\%)}} & \makecell[c]{\textbf{AlpacaEval 2.0} \\ \textbf{LC WR(\%)}} & \makecell[c]{\textbf{Arena-Hard} \\ \textbf{WR(\%)}} & \makecell[c]{\textbf{MT-Bench} \\ \textbf{Avg. Score}} & \makecell[c]{\textbf{MMLU} \\ \textbf{5-shot}} \\
        \midrule
        \multicolumn{7}{c}{\textbf{LLaMA-3.1-8B-Base}} \\
        \midrule
        Baseline & 1.17 & 8.60 & 5.24 & 4.10 & 5.60 & \textbf{64.07} \\ 
        DPO($y_t, y_w$) & 15.72 & 51.70 & 15.78 & 4.10 & 5.19 & 50.36 \\ 
        DPO($y_t, y_l$) & \textbf{25.79} & \textbf{60.64} & \textbf{24.06} & \textbf{9.04} & \textbf{5.78} & 63.17 \\ 
        \midrule \multicolumn{7}{c}{\textbf{LLaMA-3.1-8B-Instruct}} \\ \midrule
        Baseline & 33.42 & 47.94 & 41.34 & \textbf{37.10} & 7.42 & \textbf{68.80} \\ 
        DPO($y_t, y_w$) & 46.89 & \textbf{71.19} & 38.93 & 26.04 & 7.36 & 68.31 \\ 
        DPO($y_t, y_l$) & \textbf{49.07} & 68.63 & \textbf{45.82} & 29.87 & \textbf{7.55} & 68.52 \\ 
        \bottomrule
    \end{tabular}
    }
    \label{tab:reject_chose}
    \vspace{-2mm}
\end{table}

The model trained with DPO(\( y_t, y_l \)) outperforms the model trained with DPO(\( y_t, y_w \)) in most tasks. This superior performance is attributed to DPO(\( y_t, y_l \)) optimizing both textual aesthetic preferences and human preferences. When using \( y_w \) as the rejected response, the model trained with DPO(\( y_t, y_w \)) significantly outperforms the baseline model in text aesthetic scores but shows a decrease in performance in other tasks. This indicates that \( y_t \) is more aligned with aesthetic preferences, meeting our expectations.

\vspace{-3mm}
\section{Case Study}
\begin{figure}[t]
\includegraphics[width=\textwidth]{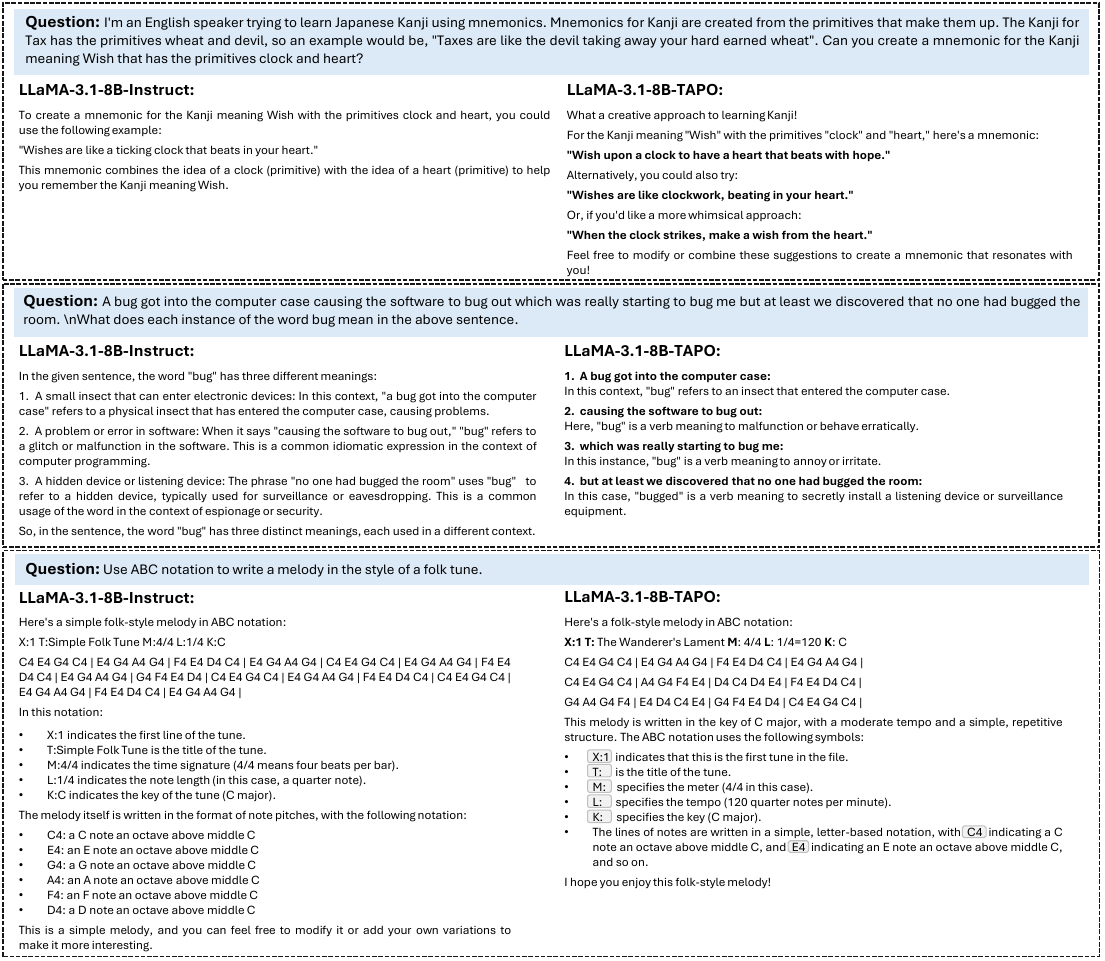}
\caption{Three cases in Arena-Hard.}
\label{fig:casestudy}
\vspace{-6mm}
\end{figure}

\vspace{-3mm}
In this section, we compare \textit{\ourbasemodel{}} and \textit{\ourmodel{}} from three cases in Arena-Hard, as shown in Figure~\ref{fig:casestudy}.

The first case (top of Figure~\ref{fig:casestudy}) examines the mnemonic for the Kanji character `Wish'. \textit{\ourmodel{}} uses bold text and line breaks effectively, making the explanation clearer and more organized. The layout, with distinct paragraph breaks and bold formatting, guides the reader seamlessly through the text.
The second case (center of Figure~\ref{fig:casestudy}) involves the term `bug'. \textit{\ourmodel{}} emphasizes each occurrence of `bug' with bold text, enhancing readability and making it easy to identify each instance. The layout is visually organized, with each explanation directly following the bolded text, ensuring a smooth flow of information and maintaining a professional appearance.
The third case (bottom of Figure~\ref{fig:casestudy}) presents a folk-style melody in ABC notation. \textit{\ourmodel{}} provides a well-organized output with logically connected ideas and adequate spacing, enhancing readability. The response is aesthetically pleasing and reader-friendly, making it easier to follow and understand the notation.
\vspace{-2mm}

\section{Conclusion}
In this paper, we conducted the first exploration of textual aesthetics in LLMs and introduced a series of techniques to enhance the aesthetic quality of LLMs outputs. First, we developed the \ourdata{} dataset, the first textual aesthetic dataset in the LLMs domain, using our specially-designed data polishing pipeline. Based on this dataset, we proposed the TAPO, which fine-tunes LLMs to improve the aesthetic quality of their outputs while preserving their core capabilities. Both qualitative and quantitative experiments validated the effectiveness of our proposed techniques. We hope our work serves as an early exploration for the textual aesthetics in LLMs and provides valuable support for researchers in the open-source community. In future work, we will continue to explore ways to collect diverse and high-quality textual aesthetics data, while designing more efficient and effective tuning techniques for aesthetic fine-tuning.

\clearpage
\vfill
\bibliography{iclr2025_conference}
\bibliographystyle{iclr2025_conference}

\clearpage
\vfill
\appendix

\section{Dataset Statistics}
\begin{figure}[t]
\begin{minipage}{0.49\linewidth}
\centering
\includegraphics[width=\textwidth]{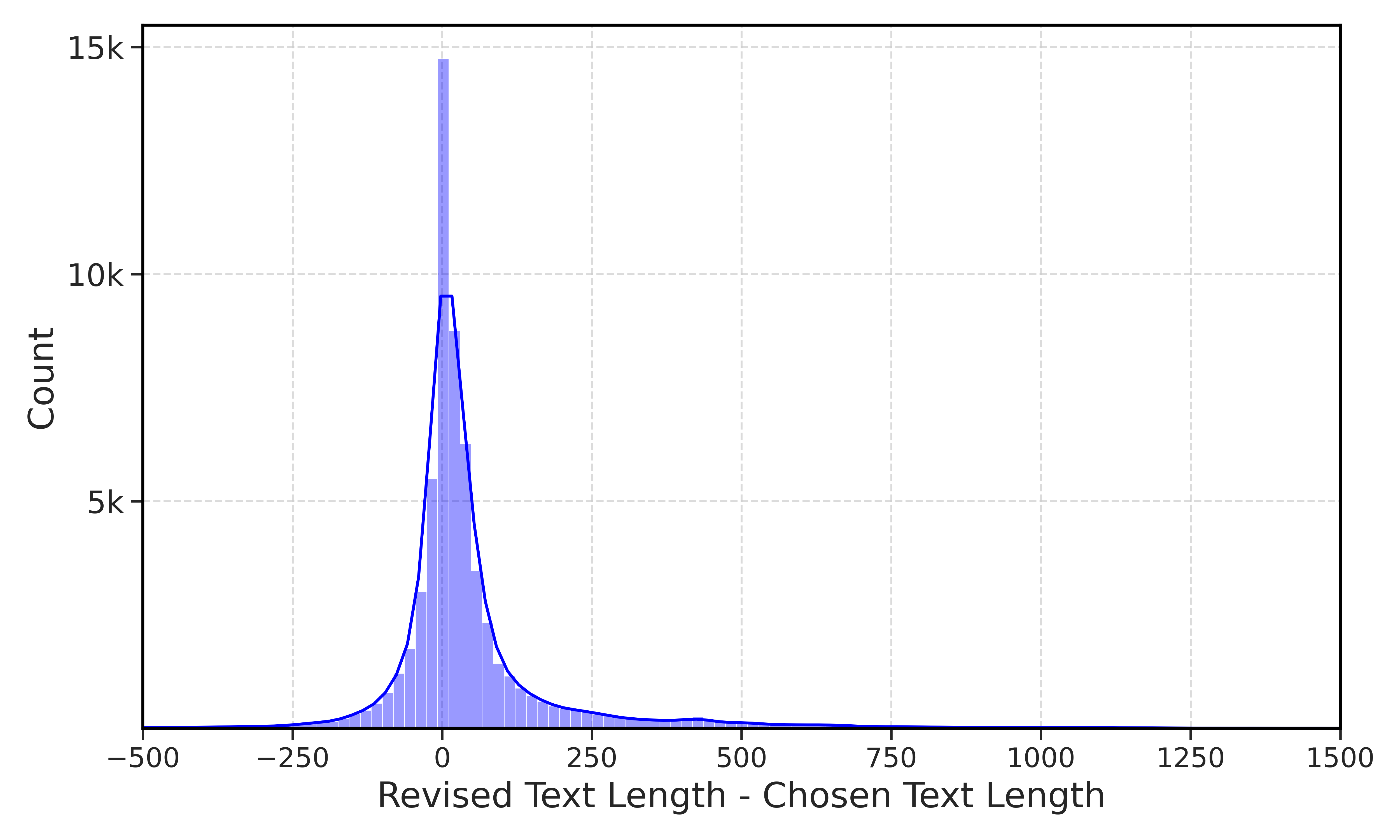}
\captionof{figure}{Distribution of length differences.} 
\label{fig:length_difference}
\end{minipage}
\hfill
\begin{minipage}{0.49\linewidth}
\centering
\setlength{\abovecaptionskip}{-0.2mm}
\includegraphics[width=\textwidth]{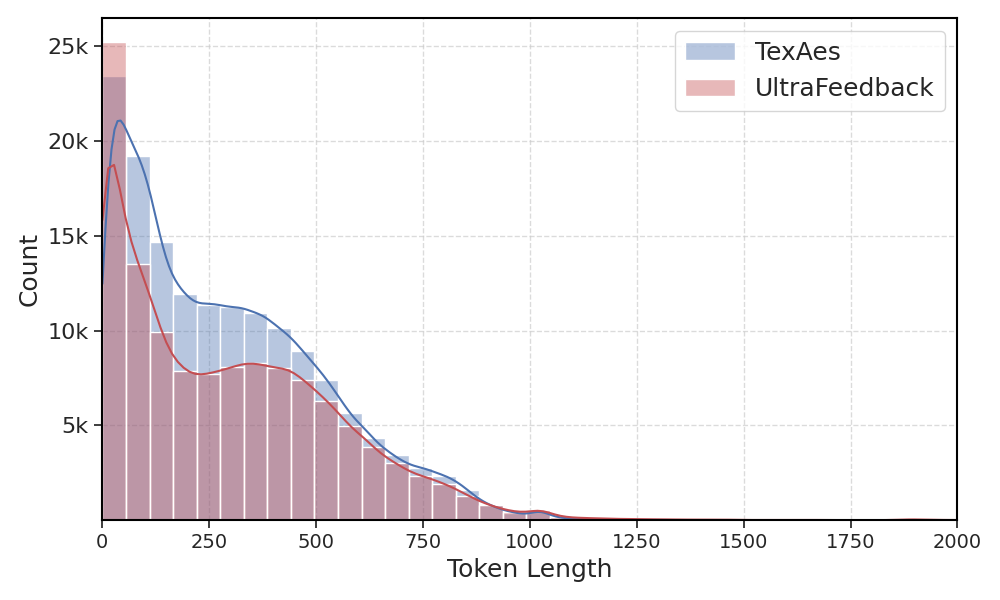}
\captionof{figure}{Comparison of token length distributions between \ourdata{} and UltraFeedback.} 
\label{fig:length_disstribution}
\end{minipage}
\vspace{-2mm}
\end{figure}

\vspace{-2mm}
\label{app:datastat}
Figure \ref{fig:length_difference} illustrates the difference in token length between the text that has undergone aesthetic polishing and the original text. The mean length difference is 49 tokens, with the 25th and 75th percentile values being -7 and 54, respectively. The maximum length difference is 2673 tokens, while the minimum length difference is -1024 tokens.


In Figure \ref{fig:length_disstribution}, we plot the length distribution of our \ourdata{}. The mean length is 293 tokens, with the 25th and 75th percentile values being 97 and 444 respectively, and the maximum length being 1408 tokens. Figure \ref{fig:length_disstribution} shows the length distribution of UltraFeedback~\citep{cui2024ultrafeedback}. The mean length is 297 tokens, with the 25th and 75th percentile values being 77 and 464 respectively, and the maximum length being 2700 tokens.

\section{Training Parameters}
\vspace{-2mm}
\label{app:param}
We present the details of the experimental settings in Table~\ref{tab:training_parameters_sft} and Table~\ref{tab:training_parameters_dpo}. For the sake of fairness in comparison, we used the same training parameters as those employed by DPO during the preference optimization stage. Our experiments are based on Llama-Factory~\citep{zheng2024llamafactory} 

\section{Mathematical Derivations}
\label{appendix:proof_equivalence}  
In this section, we prove that \(\mathcal{L}_\text{DPO}\) from Eq.~\ref{eq:loss_two_part} is equivalent to Eq.~\ref{eq:dpo_loss}. To begin, consider Eq.\ref{eq:loss_two_part}:  
  
\begin{equation}  
\begin{split}
    \mathcal{L}_\text{DPO} &= -\log \left( \frac{\exp(r_{\theta}(x, y_w))}{\sum_{i \in \{w, l\}} \exp(r_{\theta}(x, y_i))} \right)\\
    &= -\log \left( \frac{\exp(r_{\theta}(x, y_w))}{\exp(r_{\theta}(x, y_w)) + \exp(r_{\theta}(x, y_l))} \right) \\
    &= -\log \left( \frac{1}{1 + \exp(r_{\theta}(x, y_l) - r_{\theta}(x, y_w))} \right) \\
    &= -\log \sigma \left( r_{\theta}(x, y_w) - r_{\theta}(x, y_l) \right)  
\end{split}
\label{eq:derivation}
\end{equation} 

Here, \(\sigma\) denotes the sigmoid function. In Section~\ref{sec:TAPO_define}, we presented the specific expressions for \(r_{\theta}(x, y_w)\) and \(r_{\theta}(x, y_l)\):  

\begin{equation}  
    r_{\theta}(x, y_w) = \beta \log \frac{\pi_\theta(y_w \mid x)}{\pi_{\text{ref}}(y_w \mid x)}, \quad r_{\theta}(x, y_l) = \beta \log \frac{\pi_\theta(y_l \mid x)}{\pi_{\text{ref}}(y_l \mid x)}  
\label{eq:r_expressing}
\end{equation}  

By substituting Eq.~\ref{eq:r_expressing} into the Eq.~\ref{eq:derivation}, we obtain:  
  
\begin{equation}  
\quad\mathcal{L}_\text{DPO} = - \log \sigma \left( \beta \log \frac{\pi_\theta(y_w \mid x)}{\pi_{\text{ref}}(y_w \mid x)} - \beta \log \frac{\pi_\theta(y_l \mid x)}{\pi_{\text{ref}}(y_l \mid x)} \right)  
\end{equation}  
  
This shows that \(\mathcal{L}_\text{DPO}\) as defined in Eq.~\ref{eq:loss_two_part} is indeed equivalent to Eq.~\ref{eq:dpo_loss}, thus completing the proof.  

\section{Length Constraint in \ourdata{} Dataset}
\vspace{-2mm}
\label{app:data_length}
To verify whether filtering out outliers in the distribution of length differences before and after aesthetic polishing can improve the quality of \ourdata{} during its construction phase, we conducted an ablation experiment on data without length filtering. Specifically, the model was trained based on \textit{LLaMA-3.1-8B-Base} using DPO($y_t$, $y_l$), with the outcomes delineated in Table \ref{tab:length_filter}. The findings demonstrate that the performance of the model, after removing data points with excessive length deviations, significantly exceeds that of the model trained without such length filtering across all evaluation tasks. Furthermore, a statistical analysis of the output lengths generated by the model revealed that the outputs produced by the model trained with length-filtered data were not only shorter but also more concise, thereby affirming the efficacy of length filtering in text aesthetic optimization.  

\begin{table}[t]
    \centering
    \caption{Parameters for SFT training.}
    \begin{tabular}{ll}
        \toprule
        \textbf{Parameter} & \textbf{Value} \\
        \midrule
        Training Method & LoRA~\citep{hu2021loralowrankadaptationlarge} \\
        Maximum Sequence Length & 2048 \\
        Optimizer & AdamW \\
        Precision & BFloat16 \\
        Global Batch Size & 64 \\
        Maximum Learning Rate & 0.0002 \\
        Learning Rate Scheduler & Cosine with 10\% Warmup \\
        Number of Epochs & 2 \\
        \bottomrule
    \end{tabular}
    \label{tab:training_parameters_sft}
    \vspace{-4mm}
\end{table}

\begin{table}[t]
    \centering
    \caption{Parameters for \ourmethod{} training.}
    \begin{tabular}{ll}
        \toprule
        \textbf{Parameter} & \textbf{Value} \\
        \midrule
        Training Method & LoRA~\citep{hu2021loralowrankadaptationlarge} \\
        Maximum Sequence Length & 2048 \\
        Optimizer & AdamW \\
        Precision & BFloat16 \\
        Global Batch Size & 64 \\
        Maximum Learning Rate & 0.00002 \\
        Learning Rate Scheduler & Cosine with 10\% Warmup \\
        Number of Epochs & 2 \\
        Beta & 0.1 \\
        Loss Weight $w_{\text{TA}}$ & 1 \\
        Loss Weight $w_{\text{DPO}}$ & 1 \\
        \bottomrule
    \end{tabular}
    \label{tab:training_parameters_dpo}
    \vspace{-6mm}
\end{table}

\begin{table}[t]
    \centering
    \caption{Ablation study for length filter.}
    \resizebox{0.8\linewidth}{!}{
    \renewcommand\tabcolsep{4.0pt}
    \begin{tabular}{ccccccc}
    \toprule
        Length Filter & TA Text & TA Image & AlpacaEval 2.0 & Arena-Hard & MT-Bench & Avg Tokens \\ \midrule
        \ding{56} & 24.94 & 56.64 & 20.62 & 7.57 & 4.75 & 649 \\ 
        \checkmark & 25.79 & 60.64 & 24.06 & 9.04 & 5.78 & 610 \\ 
        \bottomrule
    \end{tabular}
    }
    \label{tab:length_filter}
    \vspace{-6mm}
\end{table}

\clearpage
\section{Prompt}
\vspace{-2mm}
\subsection{Aesthetics Polishing Prompt}
\label{app-prompt-polish}
\begin{tcolorbox}[
    colback=white, 
    colframe=gray, 
    title=\textbf{Prompt Template for Text Rewriting}, 
    fonttitle=\bfseries\large, 
    arc=4mm, 
    breakable
]
\textbf{System Instruction}

You are tasked with acting as a text rewriter to enhance the readability and comprehension of text generated by a Large Language Model (LLM). Your goal is to ensure the text is easy to read, easy to understand, and visually organized in a logical manner. Modifications should be reasonable and appropriate, rather than mandatory. Each element should be used judiciously to enhance readability and comprehension.

\textbf{User Instruction}

<|User Prompt|>

\{instruction\}  

<|The Start of Assistant's Answer|>

\{completion\}

<|The End of Assistant's Answer|>  

\hspace*{\fill}

Your task is to:  

1. **Analyze the LLM-generated response**:  

\hspace{2em} - Read and understand the text to grasp its context and purpose.  
    
\hspace{2em} - Carefully review the text generated by the LLM.  
    
\hspace{2em} - Evaluate its structure, formatting, and overall readability.  

2. **Determine the Need for Modification**:  

\hspace{2em} - Decide whether the text needs modification to improve its readability and comprehension.  
    
\hspace{2em} - If the text is already satisfactory, no changes are necessary.  
    
3. **Provide a Revised Version of the Text if Necessary**:  

\hspace{2em} - Make appropriate modifications to enhance the text's readability and comprehension.  
    
\hspace{2em} - Ensure the revised text maintains a consistent style and format throughout.  

\hspace*{\fill} 

**Textual Aesthetic Elements to Consider**:  

1. **Paragraph Structure**: Ensure paragraphs are of appropriate length and logically structured.  Use appropriate spacing between paragraphs.  

2. **Indentation**: Apply consistent indentation if necessary. 

3. **Headings and Subheadings**: Use headings to organize content and improve readability, but only if the content naturally lends itself to such structure.  

4. **Lists and Bullet Points**: Utilize lists to break down complex information when applicable.  

5. **Formatting for Emphasis**: Use bold or italic text to emphasize important points judiciously.  

6. **Line Spacing**: Adjust line spacing to enhance readability.  

7. **Consistency**: Maintain a consistent style throughout the document.  

8. **Visual Breaks**: Use visual breaks to separate different sections if applicable. 

9. **Blockquotes**: Use blockquotes for quotations or highlighted text.  

10. **Links**: Format hyperlinks appropriately when applicable.  

11. **Tables**: Use tables for any tabular data if required.  

12. **Whitespace and Spacing**: Ensure appropriate use of whitespace and spacing to avoid a cluttered appearance.  

\hspace*{\fill} 

**Format**:  

**Textual Aesthetic Analysis**:  

- Your analysis  

**Does it need modification**: 

- If it needs modification: [[Y]] 

- If it doesn't need modification: [[N]] 

**Revised Text**:  

- If it needs modification: <|Revised Content Start|>Your revised text<|Revised Content End|> 

- If it doesn't need modification: <|Revised Content Start|>""<|Revised Content End|>  

\hspace*{\fill} 

**Example Output**:  

**Textual Aesthetic Analysis**: 

The text is clear, well-organized, and easy to read. 

**Does it need modification**: [[N]]  

**Revised Text**:  

<|Revised Content Start|>""<|Revised Content End|> 

\end{tcolorbox}

\subsection{Text-Based TEXTUAL AESTHETICS  Scoring Prompt}
\label{app-prompt-texteval}
\begin{tcolorbox}[
    colback=white, 
    colframe=gray, 
    title=\textbf{Prompt Template for Text-Based TEXTUAL AESTHETICS Scoring}, 
    fonttitle=\bfseries\large, 
    arc=4mm, 
    breakable
]
\textbf{System Instruction}

You are an impartial judge tasked with evaluating the textual aesthetics of responses provided by two AI assistants to the user prompt displayed below. Your goal is to determine which response is more aesthetically pleasing and easier to read and understand.  

\hspace*{\fill} 

Begin your evaluation by considering the following aspects for each response:  

\hspace*{\fill} 

1. **Readability**: Is the text easy to read and understand? Are the sentences of appropriate length and complexity?  

2. **Visual Organization**: Is the text visually organized in a logical manner? Are there appropriate headings, subheadings, lists, and other formatting elements?  

3. **Consistency**: Does the text maintain a consistent style and format throughout?  

4. **Overall Structure**: Are the paragraphs well-structured and logically connected? Is there appropriate spacing between paragraphs?  

\hspace*{\fill} 

Follow these steps for your evaluation:  

1. **Analyze each response**: Carefully read and analyze both responses based on the criteria provided.  

2. **Compare both responses**: Determine which response excels in textual aesthetics considering all aspects. 

3. **Make a final decision**: Choose the response that is better in terms of textual aesthetics and justify your choice.  

\hspace*{\fill} 

You must output only one of the following choices as your final verdict with a label:  

1. Assistant A is significantly better: [[A>{}>B]]

2. Assistant A is slightly better: [[A>B]]  

3. Tie, relatively the same: [[A=B]]  

4. Assistant B is slightly better: [[B>A]]  

5. Assistant B is significantly better: [[B>{}>A]]

\hspace*{\fill} 
  
Example output: "My final verdict is Assistant A is slightly better: [[A>B]]."  

\textbf{User Instruction}

<|User Prompt|>

\{question\}

<|The Start of Assistant A's Answer|>

\{answer\_1\}

<|The End of Assistant A's Answer|>

\hspace*{\fill} 

<|The Start of Assistant B's Answer|>

\{answer\_2\}

<|The End of Assistant B's Answer|>" 
\end{tcolorbox}

\subsection{Image-Based TEXTUAL AESTHETICS  Scoring Prompt}
\label{app-prompt-imageeval}

\begin{tcolorbox}[
    colback=white, 
    colframe=gray, 
    title=\textbf{Prompt Template for Image-Based TEXTUAL AESTHETICS Scoring}, 
    fonttitle=\bfseries\large, 
    arc=4mm, 
    breakable
]
\textbf{System Instruction}

You are an impartial judge tasked with evaluating the textual and visual aesthetics of responses provided by two AI assistants to the user prompt displayed below. You will be given both the textual answers and images of the responses from each assistant. Your goal is to determine which response is more aesthetically pleasing and easier to read and understand, considering both textual and visual factors.

\hspace*{\fill} 

Evaluate each response based on the following criteria:

\hspace*{\fill} 

1. **Readability**: Is the text easy to read and understand? Are the sentences of appropriate length and complexity?  

2. **Visual Organization**: Is the text visually organized in a logical manner? Are there appropriate headings, subheadings, lists, and other formatting elements?  

3. **Consistency**: Does the text maintain a consistent style and format throughout?  

4. **Overall Structure**: Are the paragraphs well-structured and logically connected? Is there appropriate spacing between paragraphs?  

\hspace*{\fill} 

Follow these steps for your evaluation:

1. **Analyze each response**: Carefully examine both images based on the criteria provided.

2. **Compare both responses**: Determine which response excels in textual and visual aesthetics considering all aspects.

3. **Make a final decision**: Choose the response that is better in terms of textual and visual aesthetics and justify your choice.

\hspace*{\fill} 

Output your final verdict with one of the following labels:

1. Assistant A is significantly better: [[A>{}>B]]

2. Assistant A is slightly better: [[A>B]]

3. Tie, relatively the same: [[A=B]]

4. Assistant B is slightly better: [[B>A]]

5. Assistant B is significantly better: [[B>{}>A]]

\hspace*{\fill} 

Example output:   

1. Analysis of Assistant A's response:  

\hspace{2em}    - Readability: ...  
    
\hspace{2em}    - Visual Organization: ...  
    
\hspace{2em}    - Consistency: ...  
    
\hspace{2em}    - Overall Structure: ...  
    
2. Analysis of Assistant B's response: 

\hspace{2em}    - Readability: ...  
    
\hspace{2em}    - Visual Organization: ... 
    
\hspace{2em}    - Consistency: ...  
    
\hspace{2em}    - Overall Structure: ...  
    
3. Comparison:  

\hspace{2em}    - Both responses are similar in readability, but...  
    
\hspace{2em}    - Assistant A has better visual organization...  
    
\hspace{2em}    - Assistant B's consistency is...  
    
\hspace{2em}    - Overall, Assistant A/B stands out in...
    
\hspace*{\fill} 
    
My final verdict is Assistant A is slightly better: [[A>B]].

\textbf{User Instruction}

<|User Prompt|>

\{question\}

Below are two images: the first one is Assistant A's response, and the second one is Assistant B's response. Please evaluate them based on the criteria provided and give the final verdict answer.

<|The Image of Assistant A's Answer|>

\{base64\_image1\}

<|The Image of Assistant B's Answer|>

\{base64\_image2\}
\end{tcolorbox}
\clearpage

\section{Additional Examples}

\subsection{Text aesthetics judgement}
\begin{figure}[h]
\includegraphics[width=\textwidth]{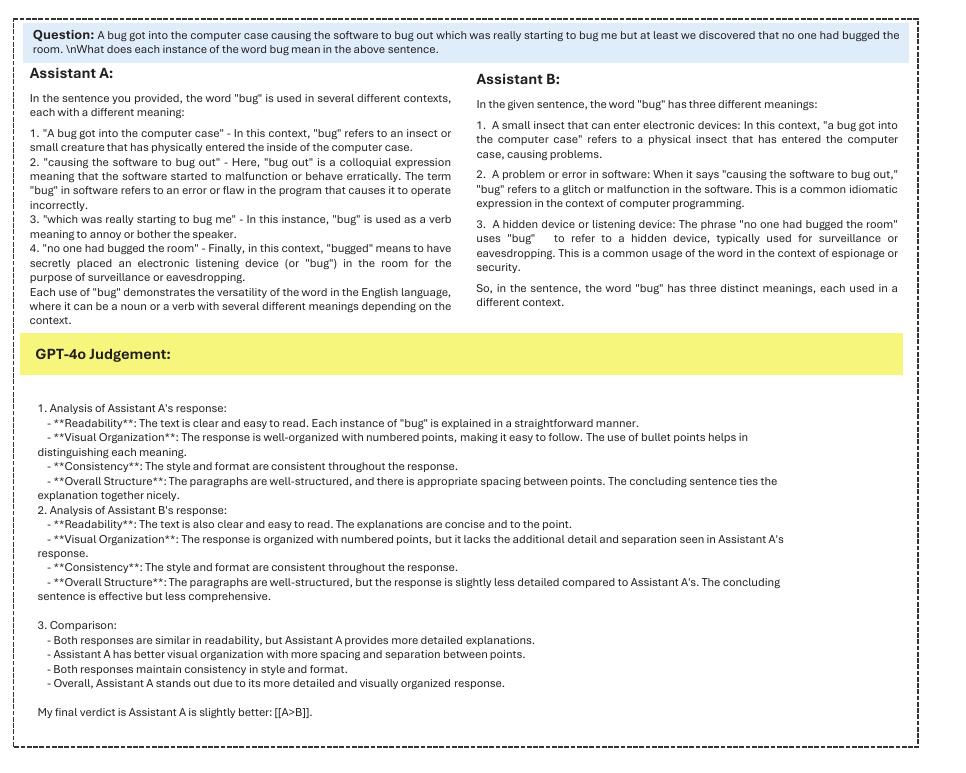}
\caption{An example of image-based GPT-4o judgement. The Assistant A is GPT-4 Turbo, and Assistant B is \ourbasemodel{}. The image demonstrates that GPT-4o can evaluate text aesthetics, showing that Assistant A's response is more visually organized and detailed compared to Assistant B's.}
\label{fig:gpt4o_judge}

\end{figure}

\end{document}